\documentclass[10pt,twocolumn,letterpaper]{article}

\usepackage{iccv}
\usepackage{times}
\usepackage{epsfig}
\usepackage{graphicx}
\usepackage{amsmath}
\usepackage{amssymb}
\usepackage{multirow}
\usepackage{multicol}

\usepackage{verbatim}
\usepackage{bbding}

\usepackage{anyfontsize}

\usepackage{makecell}
\usepackage{algorithm}
\usepackage[noend]{algpseudocode}

\makeatletter
\def\BState{\State\hskip-\ALG@thistlm}
\makeatother

\usepackage[pagebackref=true,breaklinks=true,letterpaper=true,colorlinks,bookmarks=false]{hyperref}

\iccvfinalcopy 

\def\iccvPaperID{5} 

\ificcvfinal\fi
\begin{document}

\title{Metric-based Regularization and Temporal Ensemble for Multi-task Learning using Heterogeneous Unsupervised Tasks}

\author{Dae Ha Kim, Seung Hyun Lee, and Byung Cheol Song\\
Inha University\\
100 Inha-ro, Michuhol-gu, Incheon, 22212, Republic of Korea\\
{\tt\small kdhht5022@gmail.com, lsh910703@gmail.com, bcsong@inha.ac.kr}
}

\maketitle

\begin{abstract}
One of the ways to improve the performance of a target task is to learn the transfer of abundant knowledge of a pre-trained network. However, learning of the pre-trained network requires high computation capability and large-scale labeled datasets. To mitigate the burden of large-scale labeling, learning in un/self-supervised manner can be a solution. In addition, using un-supervised multi-task learning, a generalized feature representation can be learned. However, un-supervised multi-task learning can be biased to a specific task. To overcome this problem, we propose the metric-based regularization term and temporal task ensemble (TTE) for multi-task learning. Since these two techniques prevent the entire network from learning in a state deviated to a specific task, it is possible to learn a generalized feature representation that appropriately reflects the characteristics of each task without biasing. Experimental results for three target tasks such as classification, object detection and embedding clustering prove that the TTE-based multi-task framework is more effective than the state-of-the-art (SOTA) method in improving the performance of a target task.

\end{abstract}


\section{Introduction}\label{section1}

Unsupervised learning (USL) has been used to generate pre-trained models for improving the performance of various computer vision tasks \cite{krizhevsky2012imagenet,simonyan2014very,he2016deep}. USL not only consumes less human resources because it does not need label information in learning, but also it has a merit that overfitting phenomenon is relatively less than supervised learning. In \cite{erhan2010does}, Erhan et al. analyzed in detail how the pre-trained models affect the performance of the target tasks. Their finding has recently inspired the study of deep neural networks (DNNs) based on pre-trained models.

\begin{figure}[t]
    \begin{center}
       \includegraphics[width=7.0cm, height=8.0cm]{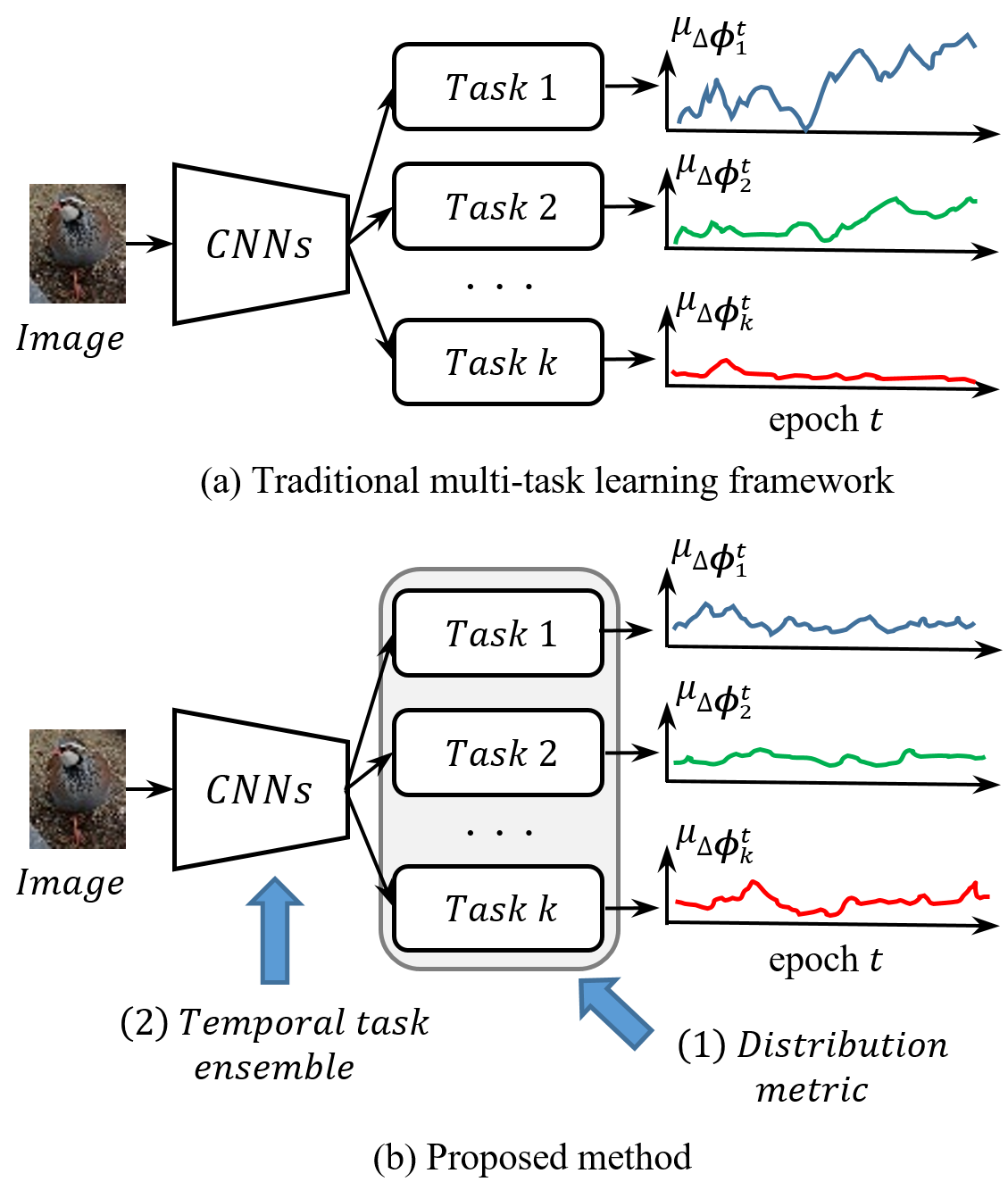}
    \end{center}
       \caption{The main concept of the proposed method. Unbalanced multi-task learning (a) can be learned evenly by using the proposed technique (b). Here, the vertical axis of the graphs indicates the degree to which task $k$ affects CNNs in epoch $t$.}
\vspace{-0.2cm}
    \label{Figure1}
\end{figure}

On the other hand, self-supervised learning (SSL) refers to learning a feature without data annotation by introducing a pretext task \cite{zhang2016colorful,noroozi2016unsupervised,mundhenk2018improvements,noroozi2018boosting}. SSL can be included in the USL category in a wide sense because it inherently learns without label information. Note that SSL has been also used to obtain pre-trained models for various target tasks such as classification and object detection \cite{doersch2017multi,noroozi2016unsupervised,mundhenk2018improvements}. 

Recently, multi-task learning (MTL), which learns a generalized feature representation using several different tasks, has been attracting attention \cite{zhang2017survey}. A lot of studies have reported that learning various supervised tasks for a source dataset outperforms learning a single task \cite{long2015learning,misra2016cross,gkioxari2015contextual,kendall2018multi}. However, supervised tasks fundamentally require labeled datasets, and as the size of the dataset increases, the labeling cost becomes significantly burdensome. To solve this problem, some MTLs composed of unsupervised tasks have been developed \cite{pinto2017learning,zamir2016generic}, but they have a limitation in obtaining synergy between different tasks because the unsupervised tasks have similar characteristics. Recently, Doersch and Zisserman proposed a new MTL using heterogeneous unsupervised tasks \cite{doersch2017multi}. However, since this method employs a simple ensemble based on a linear combination at the end of the network, it rarely derives a synergy between heterogeneous tasks either.

In order to learn more generalized feature representation than \cite{doersch2017multi}, this paper proposes a metric-based regularization term and a temporal task ensemble (TTE) for MTL using heterogeneous unsupervised tasks. The proposed regularization term is defined based on Kullback-Leibler divergence (KLD) \cite{mackay2003information} and plays a role in stabilizing latent feature maps during the learning process of heterogeneous tasks. TTE defines the difference in the degree of learning of tasks as $L_1$ norm and adaptively ensembles them.

\begin{figure*}
    \begin{center}
       \includegraphics[width=16.0cm, height=5.4cm]{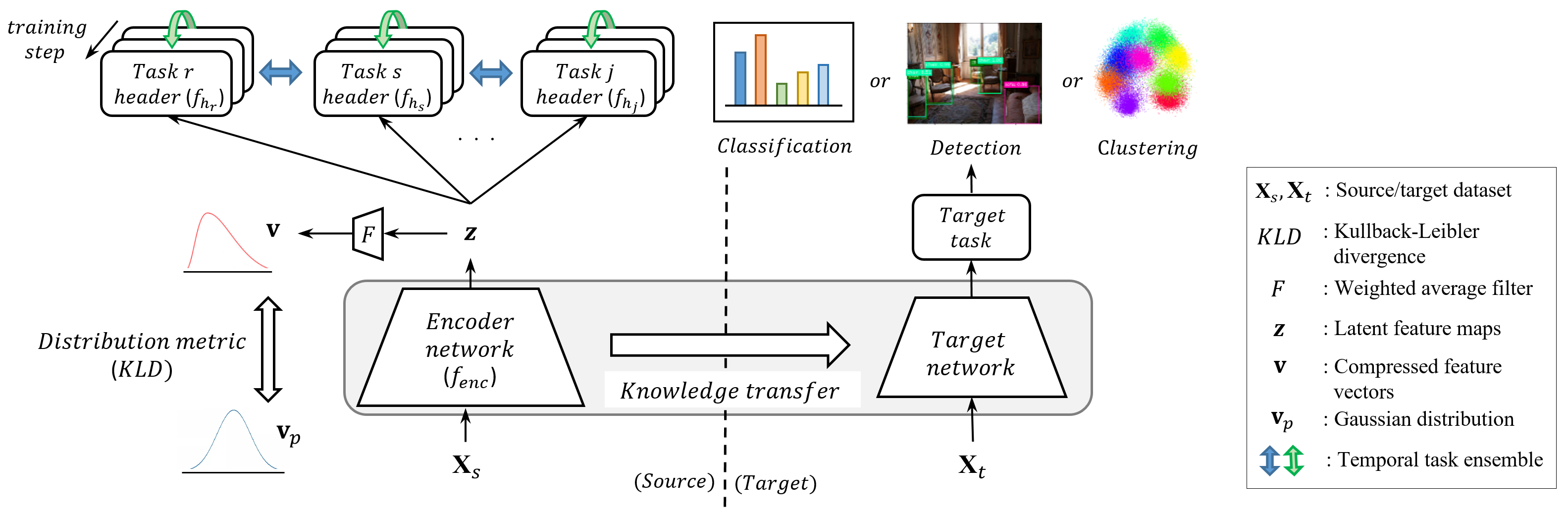}
    \end{center}
    \vspace{-0.2cm}
       \caption{Overall framework of the proposed method. The target network solves the target tasks by transferring the knowledge of the source encoder learned without a label.}
    \label{Figure2}
\end{figure*}

If the two techniques are applied to MTL, each task can be learned to affect the whole network uniformly as shown in Fig. \ref{Figure1} (b). The proposed MTL framework consists of three stages as in Fig. \ref{Figure2}. 1) MTL using an encoder-header network, 2) transferring the knowledge of an encoder network to a target network, 3) performing the target task based on the transferred knowledge.

This paper adopts four types of un-supervised tasks suitable for USL purposes: reconstruction, image segmentation, image colorization, and context-based methods. We experimentally selected the above tasks that demonstrate synergy during the TTE process \cite{ng2011sparse,xia2017w,zhang2016colorful,noroozi2016unsupervised,mundhenk2018improvements,noroozi2018boosting}.

Note that the purpose of this paper is not to suggest a new un/self-supervised task but to propose a new MTL framework to improve the performance of target tasks such as classification and object detection as in \cite{doersch2017multi}. So, this paper chooses multi-task self-supervised visual learning (MSVL) \cite{doersch2017multi}, which is the most recent MTL framework developed for the same purpose, as SOTA. Based on MSVL and the proposed framework, we derived an optimal combination of un-supervised tasks to produce an effective pre-trained network, and verified the performance of the proposed MTL for various target tasks. Our contributions are as follows:

\begin{itemize}
    \item To learn multiple heterogeneous tasks without biasing, a distribution metric-based regularization loss and TTE are proposed. This enables to learn a generalized feature representation without labeled dataset.
    \item This study is valuable in that it analyzes the relationships and synergies among various un-supervised tasks. This paper is the first to analyze the synergy between latest un/self-supervised tasks to our knowledge.
    \item This paper proposes a framework to improve the performance of a target task based on unlabeled dataset as in \cite{doersch2017multi}. It is expected to play an important role in the future works such as meta-learning and online learning.
\end{itemize}

\section{Previous Works}\label{section2}
\subsection{Unsupervised Taks}

\textbf{Reconstruction} \cite{ng2011sparse}: The reconstruction task is the most basic unsupervised task and is the key method used in the recent studies \cite{goodfellow2014generative,kingma2013auto}. Basically, the reconstruction task is used to obtain pre-trained models for learning a target task. In addition, the reconstruction task uses images of the same size as the input and output of the network, and sometimes uses noise and inputs synthetic images \cite{vincent2008extracting}.

\textbf{Image segmentation} \cite{xia2017w}: Image segmentation is a task that separates objects into meaningful regions by performing pixel-based prediction. Segmentation tasks are used for applications such as retinal and brain image analysis \cite{maninis2016deep,litjens2017survey}. Basically, we need a mask label for segmentation task. However, we can utilize existing studies that perform segmentation without labels \cite{xia2017w,kanezaki2018unsupervised}. In this paper, the basic network structure used for the segmentation task is U-Net \cite{ronneberger2015u}.

\textbf{Colorization} \cite{zhang2016colorful}: Image colorization is a task of predicting ab channels using L channel on Lab color space. Like the reconstruction and segmentation tasks, the colorization task also forms the network based on the encoder-decoder manner. Colorization task is used in various vision applications such as automatic colorization \cite{qu2006manga,larsson2016learning}.

\textbf{Context-based} \cite{noroozi2016unsupervised,mundhenk2018improvements,noroozi2018boosting}: Jigsaw puzzle \cite{noroozi2016unsupervised} is a task to divide an image into patches and to predict the position of each patch after random mixing. Jigsaw puzzle does not use encoder-decoder structure unlike the previous three tasks. Instead, it obtains the relative position information of the image patches as output values through softmax function. The jigsaw puzzle task is characterized by its superior performance as an unsupervised task despite the absence of label information. Recently, context-based learning methods based on jigsaw puzzle have been developed. Jigsaw++ \cite{noroozi2018boosting} improved learning performance by replacing up to two patches in the original puzzle with patches in a completely different image. Rotation with classification (RWC) \cite{mundhenk2018improvements} task used 2x2 patch as well as 3x3 patch and also employed patch overlapping to maximize the context information of an image. It is noteworthy that the image’s chroma blurring and yoked jitter were adopted before patch generation for improving the context learning. As a result, we construct a MTL network based on the four types of unsupervised tasks described above.

\subsection{Multi-task Learning}
MTL studies for computer vision are divided into two categories: fusion of loss functions of several tasks \cite{kendall2018multi,zamir2016generic} and fusion of information derived from CNN layers \cite{long2015learning,misra2016cross,pinto2017learning,doersch2017multi}. First, we take a look at some approaches to integrate loss functions. In \cite{zamir2016generic}, a joint framework, where the loss function of a supervised 3D task such as pose estimation and that of an unsupervised 3D task tightly coupled to the supervised 3D task are merged, was presented. And a synergy among two heterogeneous tasks was demonstrated to some extent. In \cite{kendall2018multi}, the loss function that maximizes the Gaussian probability based on task-dependent uncertainty was defined, and the weighting of each task was adjusted based on the defined loss function.

Second, a few ways to fuse information from CNN layers are described as follows. In \cite{long2015learning}, the learning ability of task-specific layers improved by designing the fully-connected (FC) layers of the network as the prior matrix. In \cite{misra2016cross}, information of convolutional layers was fused based on so-called cross-stitch units which were learned to find an optimal task combination using the activation function values of several tasks.

On the other hand, information of CNN layers was fused based on unsupervised tasks \cite{pinto2017learning,doersch2017multi}. In \cite{pinto2017learning}, a simple multi-task framework for fusing the information of CNN layers was proposed for robot grasping control purpose. In \cite{doersch2017multi}, Doersch and Zisserman presented the most successful example of combining heterogeneous unsupervised tasks. They designed the last block of the network as task-specific layers and made each layer in the block learn an unsupervised task based on the lasso ($L_1$) penalty. Also, they proposed a so-called “harmonizing” method for consistency of single feature representation. However, the above-mentioned method is limited to learning a generalized feature representation that uniformly reflects the multiple tasks. In order to overcome such a limitation, we propose a novel method which gets the effect of each task evenly.

\section{Approach}\label{section3}

\begin{figure*}
    \begin{center}
       \includegraphics[width=16.0cm, height=5.4cm]{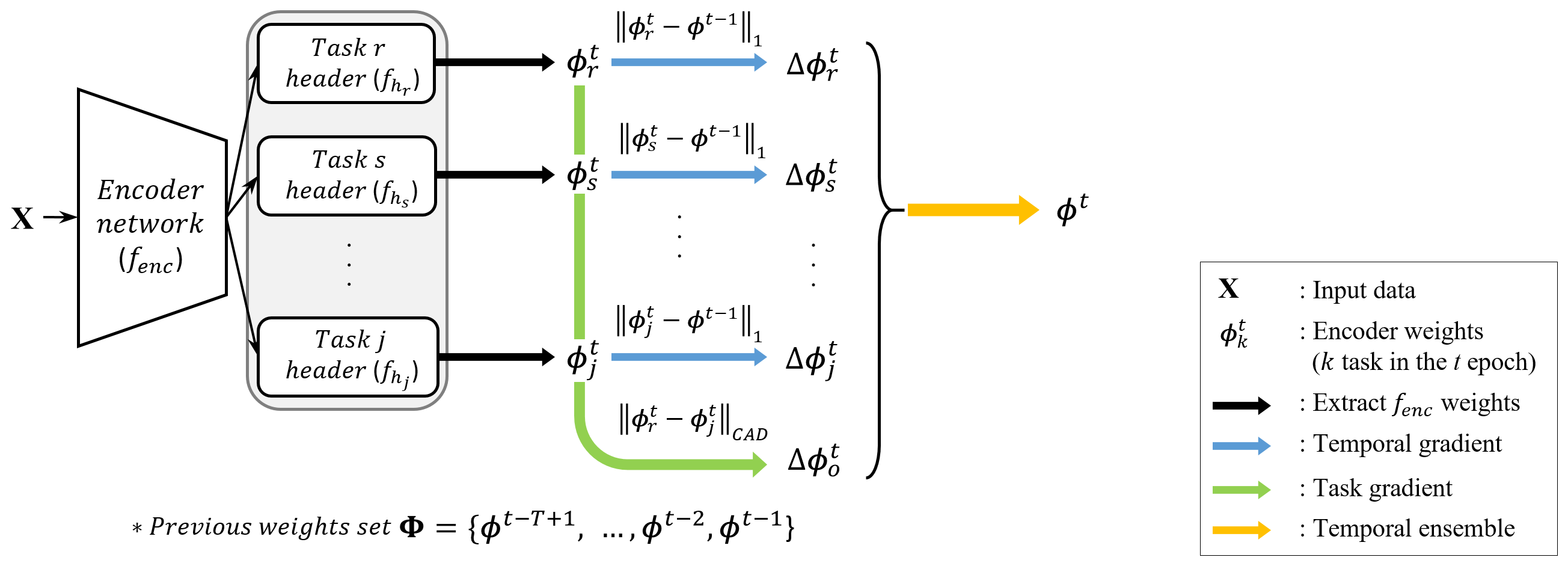}
    \end{center}
    \vspace{-0.2cm}
       \caption{The operation of TTE at epoch $t$. The order of the tasks used in this TTE process can be changed at every epoch.}
    \label{Figure3}
\end{figure*}

\subsection{Overall Architecture}\label{section3.1}

The overall architecture of the proposed method consists of two networks based on the source and target parts as shown in Fig. \ref{Figure2}. The source dataset is one for learning the pre-trained model in the source part, and the target dataset is the other for learning the target task in the target part. The source network is composed of various unsupervised tasks and has an encoder-task header structure. Each task header has an independent loss function, and the task is learned based on this loss function. The multi-task knowledge is produced by the weights of all the tasks trained in the encoder network. Finally, the multi-task knowledge is transferred to the target network. Here a target task can be classification, object detection, and embedding clustering.

Next, take a look at the learning process of the source network in detail. Let $\left\{ (\mathbf{x}, \mathbf{y}_{k}) \right\}_{i=1}^{N}$ denote a dataset consisting of $N$ samples pairs. $\mathbf{x}$ and $\mathbf{y}_k$ in a sample pair indicate an input image and the corresponding pretext of unsupervised task $k$, i.e., $k\in \left\{r, s, c, j \right\}$. '$r$', '$s$', '$c$', and '$j$' stand for reconstruction, semantic segmentation, colorization, and jigsaw puzzle, respectively. The entire operation of the proposed MTL is defined as follows.

\begin{align}\label{Eq1}
	\min_{\mathbf{W}} \sum_{k \in \left\{ r,s,c,j \right\}} \sum_{i=1}^{N} [\mathcal{L}_k \left\{ f_k(\mathbf{x}_{i};\boldsymbol{\phi},\boldsymbol{\theta}_k),\mathbf{y}_{i,k} \right\} + \Omega(\mathbf{x}_{i}; \boldsymbol{\phi})] 
\end{align}
where 
\begin{align}\label{Eq2}
	f_{k}(\mathbf{y}|\mathbf{x};{{\boldsymbol{\phi},\boldsymbol{\theta}_k}}) = f_{enc}(\mathbf{z}|\mathbf{x};\boldsymbol{\phi}) + f_{{h}_k}(\mathbf{y}|\mathbf{z};{\boldsymbol{\theta}_k})
\end{align} 
\vspace{-0.3cm}
\begin{align}\label{Eq3}
	\Omega(\mathbf{x}; \boldsymbol{\phi}) = D_{KL}(\mathbf{v}||\mathbf{v}_{p}) = \sum_{i} \mathbf{v}(i)log {\mathbf{v}(i) \over \mathbf{v}_p(i)}
\end{align}
$f_k$ indicates the network corresponding to task $k$, and consists of a common encoder network ($f_{enc}$) and a header network ($f_{{h}_k}$) as in Eq. (\ref{Eq2}). $f_k$ is learned based on $\textbf{x}$ and weights set $\mathbf{W} = \left\{ \boldsymbol{\phi}, \boldsymbol{\theta}_1, ..., \boldsymbol{\theta}_k \right\}$. $\boldsymbol{\phi}$ in $\mathbf{W}$ stands for trainable weights of the encoder network, e.g., AlexNet, VGG, and ResNet \cite{krizhevsky2012imagenet, simonyan2014very, he2016deep}. $\boldsymbol{\theta}_k$ is the trainable weights of $f_{{h}_k}$, and is configured in a manner suitable for task $k$. Also, $\mathcal{L}_k$ of Eq. (\ref{Eq1}) indicates the loss function of task $k$, which depends on $f_k$ and the corresponding pretext $\mathbf{y}_k$. 

The overall operation process is as follows. First, when $\mathbf{x}$ is input, the feed-forward process is performed to calculate $f_k$. Task $k$ is learned based on $\mathcal{L}_k$. Then, an additional constraint $\Omega$ is given to the learning process of $\mathcal{L}_k$. Because the dimension of $\mathbf{z}$ is not same as that of $\mathbf{v}_p$, distribution metric cannot be calculated directly (see Fig. \ref{Figure2}). So a transformation filter $F$, i.e., the weighted average is applied to convert a two-dimensional feature maps $\mathbf{z}$ into a one-dimensional feature vectors $\mathbf{v}$. Finally, learning proceeds towards reducing the metric distance between $\mathbf{v}$ and $\mathbf{v}_p$. Here, $\mathbf{v} \sim F(\mathbf{v}|f_{enc}(\mathbf{z}|\mathbf{x}; \boldsymbol{\phi}))$. \\
\textbf{Regularization term using distribution metric}: A regularization term   for the encoder network can be one of distribution metrics such as \textit{f}-divergence family \cite{gibbs2002choosing}. The purpose of the regularization term is to stabilize latent feature maps z during the learning process of heterogeneous tasks. We used KLD \cite{mackay2003information} as a metric for the regularization term as in Eq. (\ref{Eq3}). Verification experiments and ablation studies regarding the regularization term can be found in Section \ref{section4}.

\begin{table*}
\begin{center}

\caption{The properties of the benchmark datasets. In the domain row, S and T indicate source and target, respectively.}
\scalebox{0.9}{\begin{tabular}{|c|c c c c c c|c|}
\hline
& C10 \cite{krizhevsky2009learning} & C100 \cite{krizhevsky2009learning} & C10s & STL10 \cite{coates2011analysis} & ILSVRC2012 \cite{russakovsky2015imagenet} & Places365 \cite{zhou2014learning} & VOC0712 \cite{everingham2010pascal} \\
\hline\hline

    No. of classes & 10 & 100 & 10 & 10 & 1000 & 365 & 20 \\
    Task & \multicolumn{6}{c|}{image classification} & object detection \\
    Domain & S/T & S/T & T & T & S & S & T \\
    Samples & 60000 & 60000 & 12000 & 13000 & 1.3M & 1.8M & $\sim$20000 \\
    Image size & 32$\times$32 & 32$\times$32 & 32$\times$32 & 96$\times$96 & 224$\times$224 & 224$\times$224 & (various) \\

\hline
\end{tabular}}\label{Table1}
\end{center}
\end{table*}

\subsection{Multi-task Headers}\label{section3.2}
This section provides a detailed look at the tasks used in learning. The total loss function consists of four independent loss functions, as in Eq. (\ref{Eq4}).

\begin{align}\label{Eq4}
	\mathcal{L} := \mathcal{L}_{r} + \mathcal{L}_{s} + \mathcal{L}_{c} + \mathcal{L}_{j}
\end{align}
Here, all the loss functions except the reconstruction loss $\mathcal{L}_{r}$ are used as in \cite{xia2017w}, \cite{zhang2016colorful}, and \cite{noroozi2016unsupervised}, respectively. So we describe only $\mathcal{L}_{r}$ as follows. The first term of $\mathcal{L}_{r}$, i.e., $\mathcal{L}_{recon}$ comes from \cite{ng2011sparse}. Since the reconstruction task generally uses weak supervision than other tasks, it seldom influences the learning of the encoder network. So, in order to amplify the influence of the reconstruction task on the encoder network, a regularization term $\mathcal{L}_{reg}$ is added to $\mathcal{L}_{recon}$ as in Eq. (\ref{Eq5}).

\begin{align}\label{Eq5}
	\mathcal{L}_{r} = \mathcal{L}_{recon} + \mathcal{L}_{reg}
\end{align}
$\mathcal{L}_{reg}$ is defined as follows.

\begin{align}\label{Eq6}
	\mathcal{L}_{reg} = D_{KL}{(f_{h_{r}}{(\mathbf{y}|\mathbf{z})||\mathbf{y}_r)}} - \lambda{f_{h_{r}}{(\mathbf{y}|\mathbf{z})}\ \mathrm{log}{f_{h_{r}}{(\mathbf{y}|\mathbf{z})}}}
\end{align}
where $\mathbf{y}_r$ and $f_{{h}_r}$ indicate the label image and header of the reconstruction task, respectively. A balance factor $\lambda$ is set to $10^{-3}$. The first term of Eq. (\ref{Eq6}) is the KLD between the label image and the reconstructed image for the basic regularization effect. To design the reconstruction task to facilitate more influence on the learning of the encoder network, we employ the conditional entropy loss introduced in \cite{grandvalet2005semi} as the second term in Eq. (\ref{Eq6}). On the other hand, independent learning of the four tasks may have a limitation in causing the synergy between heterogeneous tasks, so the entire network can be learned to be biased to a specific task. Therefore, the following section provides a solution to this critical problem.

\subsection{Temporal Task Ensemble}\label{section3.3}
The core concept of TTE is to fuse the weights so that all unsupervised tasks affect evenly. In other words, TTE is targeting to suppress the entire network from being biased towards a specific task. Note that TTE does not apply to all layers of the encoder network but only to the convolution layer prior to the pooling layer. Figure \ref{Figure3} describes the detailed operation of the proposed TTE. Assume that the current training epoch is $t$. First, prepare for a set of $T$-1 encoder weights ($\Phi$) which are previously learned (see in Fig. \ref{Figure3}). Second, extract the $f_{enc}$ weights of task $k$, i.e., $\boldsymbol{\phi}_k^t$ (see black arrows in Fig. \ref{Figure3}). Next, calculate the temporal gradient $\Delta \boldsymbol{\phi}_k^t$ between $\boldsymbol{\phi}_k^t$ and $\boldsymbol{\phi}^{t-1}$ that are the encoder weights at epoch $t$-1 (see blue arrows in Fig. \ref{Figure3}).

\begin{equation}\label{Eq7}
    \Delta{\boldsymbol{\phi}_k^t} = ||\boldsymbol{\phi}_k^t - \boldsymbol{\phi}^{t-1}||_{1}
\end{equation}
where $||\cdot||_{1}$ stands for the element-wise $L_1$ distance. Note that $\Delta \boldsymbol{\phi}_k^t$ can be interpreted as the impact of task $k$ on the encoder network at epoch $t$. Also, we used the average of temporal gradients $\mu_{\Delta \boldsymbol{\phi}_k^t}$ as a measure of the impact of task $k$ (see Fig. \ref{Figure1}). Then, calculate the task gradient $\Delta \boldsymbol{\phi}_o^t$ between the encoder weights of the first task ($\boldsymbol{\phi}_r^t$ in Fig. \ref{Figure3}) and the encoder weights of the last task at epoch $t$ ($\boldsymbol{\phi}_j^t$ in Fig. \ref{Figure3}). This corresponds to a green arrow in Fig. \ref{Figure3}.

\begin{equation}\label{Eq8}
    \Delta{\boldsymbol{\phi}_o^t} = ||\boldsymbol{\phi}_r^t - \boldsymbol{\phi}_j^t||_{CAD}
\end{equation}
where $||\cdot||_{CAD}$ stands for the element-wise Canberra distance. The reason why Canberra distance is adopted here is to limit $\Delta \boldsymbol{\phi}_o^t$ to a certain range. As a result, $\Delta \boldsymbol{\phi}_o^t$ can be interpreted as a dynamic range of the weights of all tasks. Note that because the tasks are learned in an asynchronous manner during the training process, two tasks used to compute $\Delta \boldsymbol{\phi}_o^t$ in Eq. (\ref{Eq8}) may change at every training epoch. Next, based on Eqs. (\ref{Eq7}) to (\ref{Eq8}), the temporal ensemble is performed as in Eq. (\ref{Eq9}) (see a yellow arrow in Fig. \ref{Figure3}).

\begin{align}\label{Eq9}
	\boldsymbol{\phi}^t = \boldsymbol{\phi}^{t-1} + \sum_{k \in \left\{ r,s,c,j \right\}} \alpha_k^t \Delta{\boldsymbol{\phi}_k^t} + \beta^t \Delta \boldsymbol{\phi}_{o}^t
\end{align}
where $\alpha_k^t$ and $\beta^t$ indicate the adaptive coefficients at epoch $t$. The coefficients are determined adaptively by using the loss values of the current and previous epochs. Refer to \textbf{supplementary material} for the detailed calculation procedure of the coefficients. Through the ensemble process of Eq. (\ref{Eq9}), a single feature representation in which total information of all tasks is reflected can be obtained. In this process, the temporal gradient and the task gradient are tuned so that the influence of a particular task becomes not too large due to the coefficients adjusted adaptively to the loss value. In particular, the task gradient represents the dynamic range of different tasks, so coordinating this value has the same effect as allowing all tasks to be learned evenly. Finally, by taking the moving average during T time units as in Eq. (\ref{Eq10}) to prevent outliers on the time axis, we obtain the final encoder weights.

\begin{align}\label{Eq10}
	\boldsymbol{\phi} = {1 \over T} \sum_{i=0}^{T-1} \boldsymbol{\phi}^{t-i}
\end{align}
$T$ is set to 5 in this paper. Thus, the encoder weights of Eq. (\ref{Eq10}) become the multi-task knowledge for transfer to the target network. Here, we employ conventional knowledge transfer methods \cite{hinton2015distilling,yim2017gift} as mentioned in Section \ref{section3.4}.

\subsection{Knowledge Transfer Methods}\label{section3.4}
There are various studies related to transfer learning \cite{chen2018coupled,hinton2015distilling,yim2017gift,zhang2019transfer}. We make use of two conventional methods for delivering source knowledge to the target network. The first method is soft-targets \cite{hinton2015distilling} to transfer knowledge of the network output distribution. Second, we use FSP DNN \cite{yim2017gift} to transfer the flow information of the network. Note that because soft-targets do not consider the middle layer information of the network, FSP DNN may provide better performance than soft-targets. Please refer to \textbf{supplementary material} for a more detailed description of those knowledge transfer methods.

\begin{table*}
\begin{center}

\caption{The result of classification task [$\%$]. The experimental setting of the green performance was used in the ablation study in Section \ref{section4.5}. Here, Rcn, Seg, Col, jig, jig++, and RWC refer to reconstruction, segmentation, colorization, jigsaw puzzle, jigsaw++, and rotation with classification, respectively. In case of MSVL, we implemented ourselves.}
\vspace{0.1cm}
\scalebox{0.9}{\begin{tabular}{|c|c|c|c|c|}
\hline
Source dataset & Source task & Transfer method & TTE (C10s/STL10) & MSVL \cite{doersch2017multi} (C10s/STL10) \\
\hline\hline
\multicolumn{3}{|c|}{Target only} & \multicolumn{2}{c|}{61.05 / 61.19} \\
\hline
\multirow{10}{*}{ILSVRC 2012} & Jig+Col & \multirow{5}{*}{Soft-targets} & 65.43/63.83 & 61.05/56.20 \\
    & Jig+Col+Seg+Rcn & & 65.47/64.08 & 61.21/56.43 \\
    & (Jig++)+Col+Seg+Rcn & & 65.72/64.19 & \textbf{64.18}/56.79 \\
    & RWC+Col+Seg+Rcn & & \textbf{66.00}/\textbf{64.78} & 63.11/56.03 \\
    & (Jig++)+RWC+Col+Seg+Rcn & & 65.26/64.07 &63.51/\textbf{57.30} \\\cline{2-5}
& Jig+Col & \multirow{5}{*}{FSP DNN} & 69.86/66.46 & 68.97/66.23 \\
    & Jig+Col+Seg+Rcn & & \textcolor{green}{71.38}/\textcolor{green}{68.06} & 70.78/66.42 \\
    & (Jig++)+Col+Seg+Rcn & & 70.45/66.11 & 70.65/\textbf{67.75} \\
    & RWC+Col+Seg+Rcn & & \textbf{71.61}/\textbf{68.21} & \textbf{70.98}/67.43 \\
    & (Jig++)+RWC+Col+Seg+Rcn & & 71.31/67.42 & 71.08/67.40 \\[0.5ex]
    
\hline

\multirow{10}{*}{Places 365} & Jig+Col & \multirow{5}{*}{Soft-targets} & 62.87/63.12 & 61.01/62.78 \\
    & Jig+Col+Seg+Rcn & & 63.47/63.53 & 61.41/62.91 \\
    & (Jig++)+Col+Seg+Rcn & & 66.00/64.22 & 65.21/\textbf{63.98} \\
    & RWC+Col+Seg+Rcn & & \textbf{67.41}/\textbf{64.52} & \textbf{66.74}/63.74 \\
    & (Jig++)+RWC+Col+Seg+Rcn & & 65.20/64.03 &65.53/63.38\\\cline{2-5}
& Jig+Col & \multirow{5}{*}{FSP DNN} & 67.04/63.10 & 65.38/65.21 \\
    & Jig+Col+Seg+Rcn & & 67.72/63.68 & 65.60/65.53 \\
    & (Jig++)+Col+Seg+Rcn & & 67.85/64.22 & 68.67/64.57 \\
    & RWC+Col+Seg+Rcn & & \textbf{69.15}/\textbf{66.46} & \textbf{68.83}/\textbf{66.40}\\
    & (Jig++)+RWC+Col+Seg+Rcn & & 69.02/65.30 & 68.52/64.32 \\[0.5ex]

\hline
\end{tabular}}\label{Table2}
\vspace{-0.2cm}
\end{center}
\end{table*}

\begin{table*}
\begin{center}

\caption{The result of object detection task [mAP]. In case of MSVL, we implemented ourselves.}
\vspace{0.1cm}
\scalebox{0.9}{\begin{tabular}{|c|c|c|c|c|}
\hline
Source dataset & Source task & Transfer method & TTE (VOC0712) & MSVL \cite{doersch2017multi} (VOC0712) \\
\hline\hline

\multirow{5}{*}{ILSVRC 2012} & Jig+Col & \multirow{5}{*}{Fine-tune} & 61.38 & 60.41 \\
    & Jig+Col+Seg+Rcn & & 65.83 & 63.79 \\
    & (Jig++)+Col+Seg+Rcn & & 66.80 & 66.78 \\
    & RWC+Col+Seg+Rcn & & 68.71 & 67.40 \\
    & (Jig++)+RWC+Col+Seg+Rcn & & \textbf{69.35} & \textbf{68.75} \\
\hline\hline
\multicolumn{3}{|c|}{Supervised} & \multicolumn{2}{c|}{74.7} \\
\hline 
\end{tabular}}\label{Table3}
\vspace{-0.3cm}
\end{center}
\end{table*}

\begin{table*}
\begin{center}
\caption{Experimental result based on various source tasks.}
\vspace{0.1cm}
\scalebox{0.95}{\begin{tabular}{|c|c|c|c|}
\hline
Source dataset & Source task & Transfer method & Target accuracy (C10/C100) \\
\hline\hline
\multirow{9}{*}{C10 / C100} & \multicolumn{2}{c|}{Source only} & 91.55/65.37 \\\cline{2-4}
& \multicolumn{2}{c|}{Target only (baseline)} & 87.77/60.77 \\\cline{2-4}

& \multirow{3}{*}{Classification (supervised)} & Soft-targets \cite{hinton2015distilling} & 88.45/61.03 \\
& & FSP DNN \cite{yim2017gift} & 88.70/63.33 \\
& & CETL \cite{chen2018coupled} & 89.11/64.83 \\\cline{2-4}

& \multirow{2}{*}{Jig+Col+Seg+Rcn (MSVL \cite{doersch2017multi})} & Soft-targets & 88.45/62.56 \\
& & FSP DNN & 90.12/66.23 \\\cline{2-4}
& \multirow{2}{*}{Jig+Col+Seg+Rcn (TTE)} & Soft-targets & 88.73/63.07 \\
& & FSP DNN & \textbf{90.43}/\textbf{66.83} \\
\hline
\end{tabular}}\label{Table4}
\end{center}
\end{table*}

\begin{table}
\begin{center}

\caption{Performance comparison for different metrics and task difference.}
\vspace{0.1cm}
\scalebox{0.85}{\begin{tabular}{|c|c|c|}
\hline
Metric \cite{gibbs2002choosing} & \makecell{Task gradient} & \makecell{TTE (C10s/STL10)} \\
\hline\hline
    (not used) & \Checkmark & 65.40/64.33 \\
    KLD & & 68.79/65.98 \\
    KLD & \Checkmark & 71.38/68.06 \\
\hline
    Reverse KLD & \Checkmark & 69.90/66.90 \\
    JSD & \Checkmark & 68.40/65.11 \\
    Hellinger & \Checkmark & \textbf{71.48}/66.78 \\
    Jeffrey & \Checkmark & 64.48/66.22 \\
    $\chi^{2}$ & \Checkmark & 67.60/63.34 \\
    Wasserstein & \Checkmark & 70.39/\textbf{68.74} \\

\hline
\end{tabular}}\label{Table5}
\end{center}
\end{table}

\begin{table}
\centering
\caption{Performance analysis according to encoder network type.}
\vspace{0.1cm}
\scalebox{0.85}{\begin{tabular}{|c|c|c|}
\hline
\makecell{Source dataset} & \makecell{Encoder network} & \makecell{Target accuracy (C10/C100)} \\
\hline\hline
\multirow{3}{*}{\makecell{ILSVRC 2012}} & AlexNet \cite{krizhevsky2012imagenet} & 67.18/64.62 \\
& VGG16 \cite{simonyan2014very} & 69.37/66.85 \\
& ResNet50 \cite{he2016deep} & \textbf{71.38}/\textbf{68.06} \\
\hline

\multirow{3}{*}{Places 365} & AlexNet & 66.63/63.79 \\
& VGG16 & \textbf{68.70}/\textbf{65.87} \\
& ResNet50 & 67.72/63.68 \\
\hline
\end{tabular}}\label{Table6}
\end{table}

\section{Experiments}\label{section4}

\subsection{Training Configurations}\label{section4.1}
This section describes the dataset, source/target network, and learning details used in each experiment. \\
\textbf{Dataset}: Table \ref{Table1} summarizes the datasets used in all experiments. Here, C10s represents a dataset where only 20$\%$ of the CIFAR10 dataset is randomly sampled. The purpose of C10s is to verify the differentiated performance of a network by increasing the difficulty of the CIFAR10 dataset. For the evaluation of the multi-task ensemble in Section 4.2, we used ILSVRC2012 and Places365 datasets as source datasets, and C10s and STL10 as target datasets. Each image in the C10s and STL10 datasets is resized to 224$\times$224 and then input to the target task. In the knowledge distillation experiment in Section 4.3, CIFAR10/100 datasets are used as source and target datasets without any modification. \\
\textbf{Encoder/target network}: In the classification experiment in Section \ref{section4.2}, ResNet50 was used as an encoder network and the reduced AlexNet was used as a target network. The reduced AlexNet which consists of five convolutional layers and three FC layers, but the number of kernels in each layer are all reduced to about 1/2 to 1/4 of the ones in AlexNet \cite{krizhevsky2012imagenet}. Here the size of convolutional kernels is set to 3$\times$3. Then, in the object detection experiment, VGG16 was used as an encoder network and SSD300 \cite{liu2016ssd} was used as a target network. In the knowledge distillation experiment in Section \ref{section4.3}, ResNet26 was used as an encoder network and the reduced AlexNet was used as a target network. \\
\textbf{Training details}: Each numerical value in all experimental results is the average value of three trials. Each iteration is 2,000 based on the batch size of 64, and it performs up to 50 epochs. Especially, in the case of the object detection, 120 epochs. We employ stochastic gradient descent (SGD) \cite{robbins1951stochastic} with momentum 0.9 as an optimizer. Also we used the TensorFlow library for model construction as well as training.

\subsection{Evaluation of Multi-task Methods}\label{section4.2}
This section examines how the pre-trained models generated by the proposed TTE and multi-task self-supervised visual learning (MSVL) \cite{doersch2017multi} affect the performance of the target task, respectively.

First, the experimental result for the classification task is given in Table \ref{Table2}. As the number of tasks increases, the target task performance improves. Above all, TTE showed overall higher performance than MSVL. For example, using the FSP DNN method with four tasks such as RWC, Col, Seg, and Rcn, TTE improved 0.6$\%$ at C10s and 0.8$\%$ at STL10 over MSVL (see the 9-th row of Table \ref{Table2}). However, the performance improvement was relatively small when using five tasks. This implies that the source encoder is biased towards context-based tasks as Jig++ and RWC are learning together. On the other hand, the results for Places 365 show overall lower target accuracy than the results for ILSVRC2012. We are interpreting that the performance gap according to the source dataset can be due to the correlation of the dataset. Places365 consists of background/object-oriented images, while ILSVRC2012 is composed of animal/plant-oriented images. However, the target datasets CIFAR and STL10 have data characteristics more similar to ILSVRC2012 than Places365.

Second, the experimental result for the object detection task is described in Table \ref{Table3}. We started with two tasks of jigsaw puzzle and colorization, which are the most representative tasks, and increased the number of tasks to five. Table \ref{Table3} shows that the performance of both TTE and MSVL increases with the number of tasks. On average, we can observe the performance improvement of up to 8$\%$ for TTE and up to 8.3$\%$ for MSVL.

\subsection{Performance Analysis via Knowledge Distillation}\label{section4.3}
This section analyzes the performance of a target task through knowledge distillation where source and target datasets are equivalent. We experimented not only with unsupervised learning but also with supervised learning such as classification as a source task. The classification task was learned while adding the three FC layers (512-256-class dim.) after the encoder network. Soft-targets, FSP DNN, and CETL \cite{chen2018coupled} were used as transfer methods.

The first two rows of Table \ref{Table4} are the results obtained from a single network without knowledge transfer. From the third row, we show the performance of the target tasks when transferring knowledge acquired from various source tasks to the target networks.

For the transfer method of FSP DNN and the target dataset of C10, the target accuracies of the supervised task, MSVL, and TTE are 88.7$\%$, 90.1$\%$, and 90.4$\%$, respectively. The proposed method shows 0.3$\%$ better performance than MSVL. It is notable that seeing the fifth row of the last column of Table \ref{Table4}, CETL, which is known to outperform FSP DNN, shows lower performance than unsupervised methods such as MSVL and TTE. Although TTE takes about twice as long learning time and requires additional network resources than the supervised task, the fact that TTE is an unsupervised task without labeling cost and is able to provide higher performance than the supervised task can be enough to overcome the shortcomings. In addition, through this experiment, we were able to obtain an insight that the \textit{learning method of knowledge has a greater impact on performance than the transfer way of domain knowledge}.

\subsection{Deep Embedding Clustering}\label{section4.4}
This section qualitatively compares the performance of TTE and MSVL through a well-known deep embedding clustering (DEC) \cite{xie2016unsupervised} to evaluate the learning ability of the generalized feature representation. In DEC, a pre-trained encoder network and a latent distribution corresponding to the next layer, i.e., $\mathbf{z}$ in Fig. \ref{Figure2}, were used.

\begin{figure}[t]
    \begin{center}
       \includegraphics[width=6.0cm, height=12.0cm]{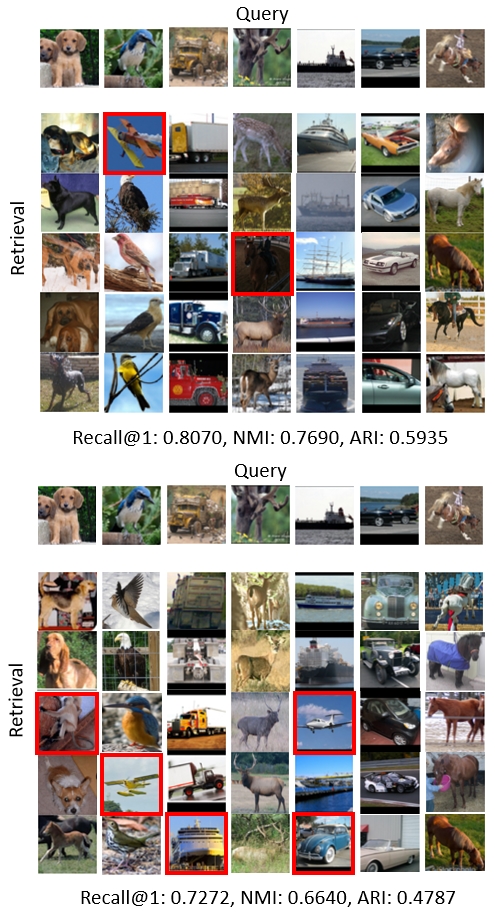}
    \end{center}
       \caption{Example retrieval results on STL10 dataset. (Upper) using encoder network trained TTE. (Below) using encoder network trained MSVL.}
\vspace{-0.3cm}
    \label{Figure4}
\end{figure}

The overall procedure is as follows. First, we constructed a cluster layer after encoder network. The cluster layer converted the output features of the encoder network into cluster label probabilities, in which a student's t-distribution was used. Next, we adjusted the cluster center based on conventional $k$-means clustering. Finally, the same learning process as \cite{xie2016unsupervised} was performed using STL10 as the target dataset.
We performed this experiment based on the encoder network of TTE and MSVL. The results are shown in Fig. \ref{Figure4} in the form of query and retrieval. TTE showed higher recall rate, normalized mutual information (NMI), and adjusted rand score (ARI) than MSVL \cite{manning2010introduction}. In addition, we can observe that more meaningful retrieval result was obtained by the proposed method.

\subsection{Ablation Study}\label{section4.5}
This section deals with the ablation study on the proposed method. All of the experiments in this section were based on the experimental setting of the green performance of Table \ref{Table2}. We analyzed the proposed method in terms of the distribution metric $\Omega$ in Section \ref{section3.1}, the application of task gradient $\Delta \boldsymbol{\phi}_o^t$ in Section \ref{section3.3}, and source encoder type, respectively.

First, we analyzed performance when replacing existing KLD with other metrics \cite{gibbs2002choosing}. As shown in Table \ref{Table5}, Hellinger and Wasserstein showed higher target task performance than KLD, which we used basically. This implies that other distribution metrics with similar constraint characteristics to KLD can be candidates. However, in case of Jeffrey and $\chi^{2}$ which are known as learning methods based on stronger constraints than KLD, their strong constraints adversely affect learning performance.

Second, Table \ref{Table5} shows the performance change according to the task gradient of Fig. \ref{Figure3}. Obviously, when the task gradient was excluded from the TTE process, the performance decreased, but this was less than when the distribution metric was not used at all.

Third, we analyze the target accuracy according to the type of encoder network. Seeing Table \ref{Table6}, ResNet50 has the highest target accuracy in the ILSVRC2012 and VGG16 has the best in the Places365. Unlike our expectation that more complex encoder network would be more beneficial for MTL, we can observe that performance were not significantly affected by the type of encoder network. As a result, all options used in the experiment are directly related to the performance of the target task, and the distribution metric has the greatest effect on performance.

\section{Conclusion}\label{section5}
In this paper, we proposed two methods that are metric-based regularization and TTE for obtaining generalized feature representation using unsupervised tasks. As a result, metric-based regularization loss and TTE make it possible to learn a pre-trained model that accurately reflects data characteristics even for large datasets such as ILSVRC2012. However, the proposed method uses the basic weighted sum form to fuse task information. Therefore, our future study will suggest a new type of task ensemble technique that can show better synergy between tasks.

\subsection*{\textbf{\fontsize{12}{12}\selectfont Acknowledgements}}
This work was supported by National Research Foundation of Korea Grant funded by the Korean Government (2016R1A2B4007353) and the Industrial Technology Innovation Program funded By the Ministry of Trade, industry $\&$ Energy (MI, Korea) [10073154, Development of human-friendly human-robot interaction technologies using human internal emotional states].

{\small
\bibliographystyle{ieee}
\bibliography{egbib}
}

\clearpage

{\LARGE\bfseries Supplementary Materials}
\vspace{0.5cm}

\subsection*{\textbf{\fontsize{12}{12}\selectfont 1. TTE coefficients}}
\label{sec:intro}
The two coefficients used during temporal task ensemble (TTE), i.e., $\alpha_k^t$ and $\beta^t$ are adaptively determined based on the loss values of the current epoch as well as the previous epoch (see. Eq. (\ref{Eq1})-(\ref{Eq3})).

\begin{align}\label{Eq1}
	\alpha_k^t = \alpha_k^{t-1} / (1+m)
\end{align}
\begin{align}\label{Eq2}
	\beta^t = \beta^{t-1} / (1+n)
\end{align} \\
where

\begin{align}\label{Eq3}
	m = [ \mathcal{L}_k^t - \mathcal{L}_k^{t-1} ]_{-},\ \ n = [ \mathcal{L}^t - \mathcal{L}^{t-1} ]_{-}
\end{align} \\
Note that $k \in \left\{ r,s,c,j \right\}$ and $t$ is the current epoch. $[\cdot]_{-}$ indicates $\mathrm{min}(0,\cdot)$. $\mathcal{L}_k^t$ is the loss function corresponding to task $k$ at epoch $t$, and $\mathcal{L}^t$ is the total loss function at epoch $t$. The initial values are set as follows: $\alpha_r^0=0.4$, $\alpha_s^0,\alpha_c^0,\alpha_j^0=0.2$, and $\beta^0=5\times 10^{-3}$. The reason for designing $m$ and $n$ as shown in Eq. (\ref{Eq3}) is to limit the influence of temporal ensemble process when the loss gap between the current epoch and the previous epoch becomes large.

\subsection*{\textbf{\fontsize{12}{12}\selectfont 2. Knowledge Transfer Methods}}
\label{sec:related}
This section discusses in detail how to transfer domain knowledge. Basically we use the typical knowledge transfer methods ~\cite{hinton2015distilling, yim2017gift}. However, since we use the unsupervised learning scheme in source domain, we add some variations to the existing knowledge transfer methods.

\subsection*{\textbf{\fontsize{12}{12}\selectfont 2.1. Details of Soft-targets}}
We first introduce soft-targets. Soft-targets are a method of transferring domain knowledge of network output distribution. This makes it easy to deliver domain knowledge, no matter how complex the source and target networks are. We perform knowledge transfer based on the output of the encoder network, i.e., the latent distribution $\mathbf{z}$.

However, we face with a problem that the latent distribution cannot be used directly. This is because we use the unsupervised task and cannot directly obtain the class distribution of the encoder network. Therefore, we construct an additional network with latent distribution $\mathbf{z}$ as input and class distribution of the encoder network $\mathbf{z}_c$ as output, like $f_{add}(\mathbf{z}_c|\mathbf{z}; \boldsymbol{\theta}_{add})$. Here $f_{add}$ is designed based on a fully-connected layers and consists of 128, 256, and 10 dimensions, respectively. For example, the last 10 indicates the dimension of $\mathbf{z}_c$. As a result, we connect $f_{add}$ to the encoder network $f_{enc}$ and design a final network like Eq. (\ref{Eq4}).

\begin{align}\label{Eq4}
	f_{st}(\mathbf{z}_c|\mathbf{x};\boldsymbol{\phi},\boldsymbol{\theta}_{add}) = f_{enc}(\mathbf{z}|\mathbf{x};\boldsymbol{\phi}) + f_{add}(\mathbf{z}_c|\mathbf{z}; \boldsymbol{\theta}_{add})
\end{align} \\
Thus, knowledge transfer is performed through $f_{st}$ as in \cite{hinton2015distilling}. Note that the parameters of the encoder network are not learned and only the parameters of the fully-connected layer are learned. This is because when the domain knowledge is transferred using \cite{hinton2015distilling}, the encoder network must transfer the information of the source domain completely.

\subsection*{\textbf{\fontsize{12}{12}\selectfont 2.2. Details of FSP DNN}}
Next, we depict how to transfer layer information of the network. We perform knowledge transfer using the flow of the solution process (FSP) matrix \cite{yim2017gift}. The FSP matrix is produced based on feature maps between two consecutive layers in the source/target model. Therefore, we form a total of five FSP matrices based on the layers in which the number of convolution filters change in the encoder network. In the target network as well, a total of five FSP matrices are generated. Then, based on the five matrix pairs, knowledge transfer is performed as in \cite{yim2017gift}.

\end{document}